# Predicting Cognitive Decline with Deep Learning of Brain Metabolism and Amyloid Imaging

Hongyoon Choi, M.D., Ph.D.[1,†], Kyong Hwan Jin, Ph.D. [2, 3, †], for the Alzheimer's Disease Neuroimaging Initiative[*]

[1]Department of Nuclear Medicine, Cheonan Public Health Center, Chungnam, Republic of Korea; [2]Department of Bio and Brain engineering, Korea Advanced Institute of Science and Technology, Daejeon, Republic of Korea; [3]Biomedical Imaging Group, École Polytechnique Fédérale de Lausanne (EPFL), Lausanne, Switzerland

[†]These authors contributed equally to this work

[*]Data used in preparation of this article were obtained from the Alzheimer's Disease Neuroimaging Initiative (ADNI) database (adni.loni.usc.edu). As such, the investigators within the ADNI contributed to the design and implementation of ADNI and/or provided data but did not participate in analysis or writing of this report. A complete listing of ADNI investigators can be found at: http://adni.loni.usc.edu/wp-content/uploads/how_to_apply/ADNI_Acknowledgement_List.pdf

Hongyoon Choi, M.D.,Ph.D.

Tel: 82-41-521-2572, Fax: 82-41-521-2587, E-mail: chy1000@gmail.com

Kyong Hwan Jin, Ph.D.

Tel : +41-78-684-90-07, E-mail : kyonghwan.jin@gmail.com

[Cover Title] Deep learning for predicting cognitive decline

# Abstract

**OBJECTIVE** We aimed to develop an automatic image interpretation system based on a deep convolutional neural network (CNN) which can accurately predict future cognitive decline in mild cognitive impairment (MCI) patients using flurodeoxyglucose and florbetapir positron emission tomography (PET).

**METHODS** PET images of 139 patients with AD, 171 patients with MCI and 182 normal subjects obtained from Alzheimer's Disease Neuroimaging Initiative database were used. Deep CNN was trained using PET images of AD and normal controls. Feature extraction and spatial normalization which have been routinely used in conventional quantitative analyses were unnecessary for our approach. Cognitive outcome of MCI subjects was predicted using this network. The prediction accuracy of the conversion of mild cognitive impairment to AD was compared with conventional feature-based quantification approach. Output variables of the network were correlated with the longitudinal change of cognitive measurements.

**RESULTS** Accuracy of prediction (84.2%) for conversion to AD in MCI patients outperformed conventional feature-based quantification approaches. ROC analyses revealed that performance of CNN-based approach was significantly higher than that of the conventional quantification methods (p < 0.05). Output scores of the network were strongly correlated with the longitudinal change in cognitive measurements (p < 0.05).

**CONCLUSIONS** Deep CNN could accurately predict future cognitive declines in MCI patients by automatically interpreting PET images. These results show the feasibility of deep learning as a practical tool for developing predictive imaging biomarker.

## Abbreviations

AD        Alzheimer's disease

MCI       Mild cognitive impairment

FDG       $^{18}$F-fluorodeoxyglucose

AV-45     $^{18}$F-florbetapir

CNN       Convolutional neural network

ADNI      Alzheimer's Disease Neuroimaging Initiative

NC        Normal control

CDR-SB        Clinical Dementia Rating Sum of Boxes

ADAS-Cog        Alzheimer's Disease Assessment Scale-Cognitive Subscale 13-item

FAQ       Functional Activities Questionnaire

MMSE   Mini-Metal Status Examination

ROC       Receiver operating characteristic

## Introduction

Recent treatment strategies for Alzheimer's disease (AD) are aimed at slowing cognitive decline and are focused on the pre-dementia stage which includes mild cognitive impairment (MCI) [1]. However, as the etiology of MCI is heterogeneous, MCI patients show different rates of cognitive decline, and even some never convert to AD [2]. Thus, it is a matter of utmost importance to identify the patients with MCI who would benefit from treatment. So far, several studies have investigated a number of imaging biomarkers that can predict whether a patient with MCI will convert to AD. They include brain metabolism and amyloid load measured by [18]F-fluorodeoxyglucose (FDG) and [18]F-florbetapir (AV-45) positron emission tomography (PET), respectively [3-6]. Previous studies have used quantitative parameters or visual assessment of the brain images for predicting MCI patients who would covert to AD. However, visual analysis cannot provide quantitative and objective data and quantitative analyses commonly require complicated processing [7-10].

In this study, we developed a deep convolutional neural network (CNN) based method, a type of deep learning, for the prediction of cognitive decline. Recent advances in CNN have dramatically improved image recognition field [11]. We applied CNN to FDG and AV-45 PET images to predict cognitive decline in MCI patients. Our automated method was designed to discriminate patients groups classified according to the cognitive outcome with minimized image processing. In addition, it could provide a quantitative biomarker combining both the FDG and AV-45 PET information. We showed that the CNN-based biomarker strongly correlated with future cognitive decline.

## Methods

### Subjects

The data used in this study included subjects recruited in Alzheimer's Disease Neuroimaging Initiative-II (ADNI-2) with available baseline data on FDG and AV-45 PET (http://adni.loni.usc.edu). The ADNI was launched in 2003 as a public-private partnership, led by Principal Investigator Michael W. Weiner, MD, VA Medical Center and University of California San Francisco. Subjects have been recruited from over 50 sites across the US and Canada. The primary purpose of ADNI has been to test whether serial MRI, PET, other biological markers, and clinical and neuropsychological assessment can be combined to measure the progression of MCI and early AD. For up-to-date information, see http://www.adni-info.org. Written informed consent to cognitive testing and neuroimaging prior to participation was obtained, approved by the institutional review boards of all participating institutions.

For diagnostic classification and learning the CNN, we firstly selected patients with AD and healthy subjects who had baseline FDG and AV-45 PET scans. We also selected MCI patients who had baseline FDG and AV-45 PET scans and 3-year follow-up clinical evaluation. This resulted in 182 normal controls (NCs), 139 AD patients and 171 MCI subjects. Based on whether the MCI patients would convert to AD within 3 years, MCI patients were grouped as MCI converters and nonconverters.

Cognitive function of the subjects was evaluated using Clinical Dementia Rating Sum of Boxes (CDR-SB), Alzheimer's Disease Assessment Scale-Cognitive Subscale 13-item (ADAS-Cog), Functional Activities Questionnaire (FAQ), and Mini-Metal Status Examination (MMSE). As we interested in whether CNN-based baseline PET biomarkers would be associated with longitudinal cognitive decline, longitudinal changes of cognitive measurements were also assessed at 1 year and 3 years after the baseline study.

**FDG PET and AV-45 PET**

All the PET images were downloaded from ADNI database at the most advanced preprocessing stage. FDG PET images were acquired 30 to 60 min and AV-45 PET images were acquired 50 to 70 min after the injection. The FDG and AV-45 PET images were co-registered to each other, averaged across the time frames, standardized to have same voxel size (1.5 x 1.5 x 1.5 mm). PET images were acquired in the 57 sites participating in ADNI, scanner-specific smoothing was additionally applied [12]. As the processing did not include nonlinear spatial warping, size and shape of the brains were not changed after the preprocessing. These preprocessed images could be downloaded from ADNI database and we used them for deep CNN training and testing as they are.

**Study design**

The main purpose of this paper was to develop a deep CNN-based method for prediction of cognitive decline and selection of subjects who would eventually convert to AD. Before the testing of MCI conversion, the deep CNN was trained using PET images of AD and NC subjects. For discrimination between MCI converter and nonconverter, the pre-trained network trained by AD/NC data was directly used as the imaging features of MCI converter would be similar with those of AD. The nodes of the output layer were only reassigned to MCI converter and nonconverter. All the PET images of MCI subjects were tested whether they would convert to AD or not. Therefore, our deep CNN was a classifier independent from the training data to discriminate between MCI converter and nonconverter (**figure 1**). In addition, we also obtained a quantitative score for MCI converter. The quantitative value of the final output of the network was defined as ConvScore, a score that indicates how close inputted baseline images are to AD. The score can be expected to be utilized for a predictive biomarker.

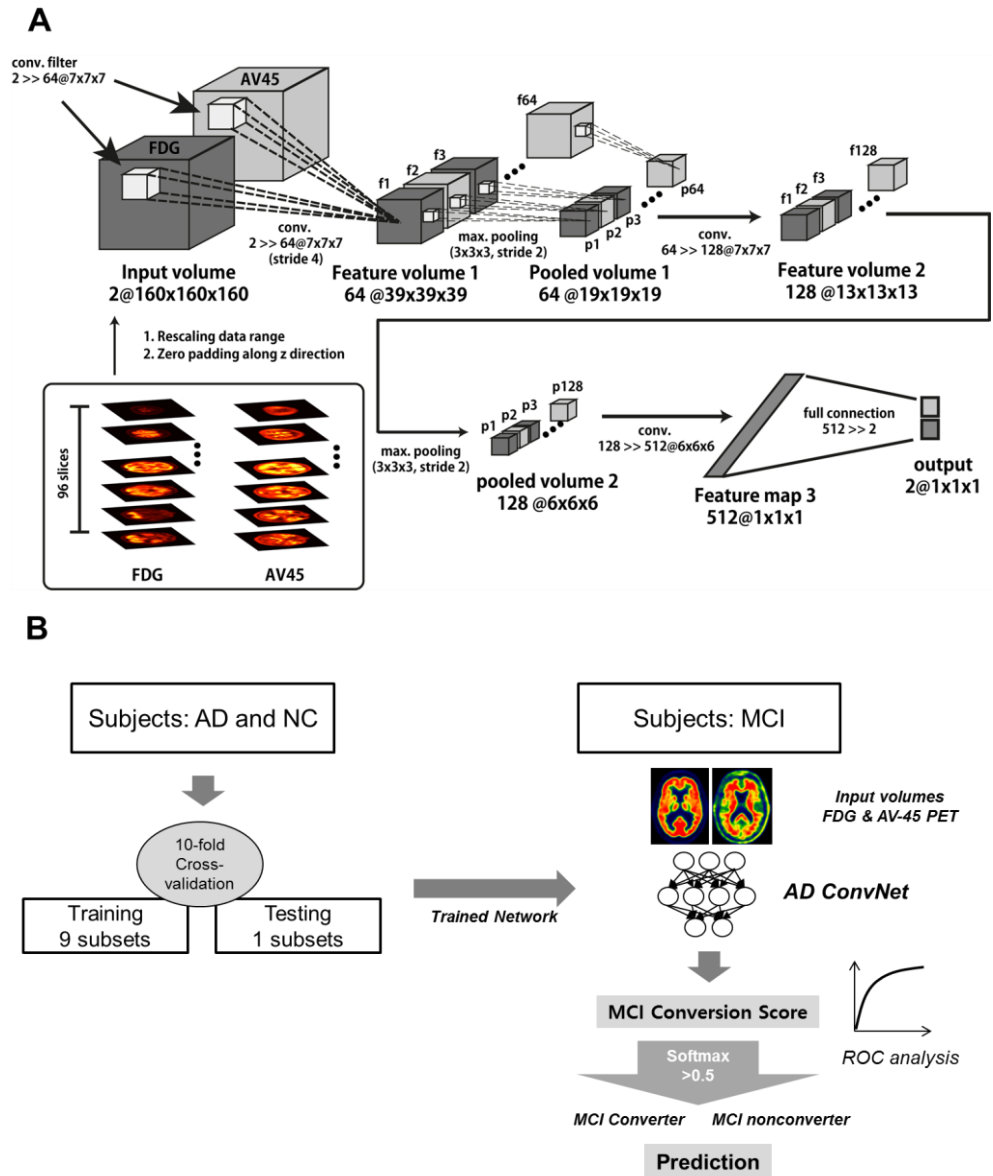

**Figure 1** Framework for predicting cognitive decline in mild cognitive impairment patients. (A) Deep convolutional neural network architecture is applied to the two PET images, FDG and florbetapir (AV-45). Each layer, features can be extracted by 3-dimensional convolution and activation (ReLU) function. Multilayer convolutions yield 1-dimensional output and the last layer have two nodes, which correspond to Alzheimer's disease (AD) and normal control (NC). (B) Deep CNN was trained from PET data of AD and NC. 10-fold cross validation was used. After the training, the trained network was directly used for the classification between mild cognitive impairment (MCI) converter and nonconverter. Thus, the network was independent from the PET data of MCI patients. Prediction accuracy for MCI conversion was assessed. Receiver operating characteristic (ROC) analysis was also performed.

**Preparation for 3-dimensional deep convolutional neural network**

Recently, classifications in image data using deep convolutional neural network (CNN) have demonstrated outstanding performances compared with previous machine learning approaches. The main reason for superior performance is originated from shift invariant property of CNN and multi-layered interpretation, which is closely related with actual human visual perception. In particular, the shift invariant property originated from the convolution operator can efficiently detect target objects distributed on overall spatial space. Deep CNN can be easily extended into 3 dimensional volume data when including natural image features such as video processing [11]. In this paper, we aimed to classify PET brain dataset using 3D CNN model. Because 3D CNN has robust shift invariant property, the detection of specific features in the 3 dimensional spaces can be effectively performed. Furthermore, in our 3D CNN model, we used two different kinds of PET data as multi-channel data (2 channels) to exploit diverse features from FDG and AV-45.

The one of the most important part of our approach is simplified pre-processing. After the aforementioned preprocessing including co-registration of two modalities, voxel-size standardization and scanner-specific smoothing, we simply rescaled the image data. The rescaling included two procedures: (1) Each modality (FDG and AV-45) is rescaled by the range from 0 to 255, and then (2) mean scalar value from a group of each modality is subtracted from all subjects. After two-step pre-processing, 3D volume data from FDG and AV-45 ( 160 x 160 x 96 volume) are concatenated as a 4D volume (160 x 160 x 96 x 2) to be used for input argument of the network.

**Network architecture**

The architecture of our network is as follows: we constructed three convolution layers followed by both maximum pooling and ReLU, and one fully connected layer. The sizes of 3-D convolution filter are (first layer: $7 \times 7 \times 7$, second layer: $7 \times 7 \times 7$, third layer: $6 \times 6 \times 6$), respectively. The number of feature maps is increased as 2, 64, 128, and 512 proportional to the depth of layer.

The relationship of 3-D convolution related with feature maps between input and output of CNN can be represented as

$$O[x, y, z, f] = C_i(I, s) = \sum_{x'=1}^{N_x} \sum_{y'=1}^{N_y} \sum_{z'=1}^{N_z} \sum_{f'=1}^{N_f} I[x'-x, y'-y, z'-z, f']s[x', y', z', f', f],$$

where $i$ is the stride number which is sliding unit of convolution operation, $I \in R^{N_x \times N_y \times N_z \times N_f}$ is an input volume for the specific convolution layer, $O \in R^{(N_x-W_x+1) \times (N_y-W_y+1) \times (N_z-W_z+1) \times N_f'}$ is an output volume from the specific convolution layer, $s \in R^{W_x \times W_y \times W_z \times N_f \times N_f'}$ is a filter bank for 3D volume with multiple feature maps, $(W_x, W_y, W_z)$ are the spatial dimensions $(x, y, z)$ of filter bank, $(N_x, N_y, N_z)$ are the spatial dimensions of input volume, $(N_f, N_f')$ are the number of feature maps before convolutional layer and after convolutional layer, respectively. For example, in the case of our first convolution layer, let we define $\hat{I}$ from a subject has $\hat{I} \in R^{160 \times 160 \times 160 \times 2}$ (increased dimension along z-axis by zero-padding), and the output $y \in R^{39 \times 39 \times 39 \times 64}$ from convolution layer (stride = 4) consisting of $s \in R^{7 \times 7 \times 7 \times 2 \times 64}$ filter (dimension of x-axis, dimension of y-axis, dimension of z-axis, # of feature maps in input, # of feature maps in output). After convolution operation, ReLU and max pooling operations will be performed on the output of convolution given as

$$O[x, y, z, f] = R(I) = \max(0, I[x, y, z, f])$$

$$O[x, y, z, f] = M_s(I) = \max_{-\lfloor s/2 \rfloor \leq i \leq -\lfloor s/2 \rfloor, -\lfloor s/2 \rfloor \leq j \leq -\lfloor s/2 \rfloor, -\lfloor s/2 \rfloor \leq k \leq -\lfloor s/2 \rfloor} I[x+i, y+j, z+k, f]$$

where $R$ is ReLU operator, $M_s$ is max pooling operator. In case of max pooling operator, the stride operation (interleaved sliding of positions) could be accompanied for reduction of network size. In practical point of views, the ReLU keeps positive input values whereas negative input values are changed into 0s. The max pooling operation chooses the maximum value from pre-defined region of 3-D box, and puts this value into reduced sized feature map.

Using aforementioned operations, we can construct efficient multi-layer CNN described as

$$O(I; \boldsymbol{s}) = \boldsymbol{C}(I; \boldsymbol{s}) = Soft(FC(R \circ C_3(M_3 \circ R \circ C_2(M_3 \circ R \circ C_1(I, s_1), s_2), s_3), s_4))$$

where $\circ$ is the sequential composition of operators, $FC$ is a fully connected layer, $\boldsymbol{s}$ is an augmented volume of each level's filter bank ($\boldsymbol{s} = (s_1, s_2, s_3, s_4)$) , $Soft$ is the softmax function. For the training phase, we choose supervised learning, so the functionals ($y(x)$) from the CNN are set to be one for AD and two for NC. The total loss function, $L(\boldsymbol{s})$, to be minimized is described as

$$L(\boldsymbol{s}) = \frac{1}{N_s} \sum_{i=1}^{N_s} l\big(O_i, \boldsymbol{C}(I_i; \boldsymbol{s})\big),$$

where $l(O_i, \boldsymbol{d_i})$ is a loss function between true label ($O_i$) with respect to input ($\boldsymbol{d_i}$), and $N_s$ is the total number of training data set.

This supervised learning are conducted by stochastic gradient descent (SGD) algorithm [13], the source code of which is distributed by MatConvNet (Version 1.0-beta 16) [14]. The detailed explanation about backpropagation and gradient descent can be referred by reference [14]. Training of the network was performed by imaging data of AD and NC. We choose the 10 fold cross-validation for supervised learning, thus, 10% of the dataset were used for the validation. The network was trained for 50 epochs. The momentum parameter for SGD was set to 0.9. The learning rate was initially $1 \times 10^{-5}$ and logarithmically decreased to $1 \times 10^{-7}$.

Before passing through the last layer, softmax function, values of two nodes indicate scores for AD and NC (or MCI converter and nonconverter), respectively. The quantitative value of 'AD node' was defined as ConvScore, a score that indicates how input PET data are close to AD or MCI converter.

**Training and validation of classification between AD and NC**

Our proposed network was learned from the image data of AD and NC. Ten-fold cross-validation was implemented to evaluate the performance and optimize the network parameters. Subjects were randomly divided into 10 subsets including a subset for testing and remaining nine subsets for training. Training and evaluating process were repeated and classification performance was evaluated using results of testing subsets. For training, convolutional layers and fully-connected layers were initialized with random weights. To increase the performance for test sets, training data were augmented as deep CNN-based 2-dimensional image recognition tasks [15]. Considering anatomical symmetry of the brain, the images were left-right flipped to augment the training set. As 90% of AD/NC PET image data were included in the training set for each fold, 289 x 2 (or 288 x 2) image pairs were used for the training. The remained 32 (or 31) image pairs were included in the validation set.

Sensitivity, specificity and accuracy of the classification were calculated by results from all validation subsets. After the softmax function, if the probability score of AD was higher than 0.5, predictive diagnosis was 'AD'. In addition, receiver operating characteristic (ROC) analysis was performed using ConvScore, output variables of 'AD node'.

**Prediction of cognitive decline in MCI subjects**

Using the network trained by PET images of AD and NC, PET images of MCI subjects were tested whether they would convert to AD or not. The labels of the output nodes of the CNN, AD and NC, were changed to MCI converter and nonconverter, as aforementioned above. As the classification of AD/NC, we classified a patient as a predictive MCI converter if the probability of the network output layer was higher than 0.5. Sensitivity, specificity and accuracy were measured and ROC analysis was also performed using ConvScore.

ConvScore was correlated with the longitudinal changes of cognitive measurements including CDR-SB, ADAS-Cog, FAQ and MMSE. Cognitive measurements of 1-year and 3-year follow-up visits were compared with those of baseline studies to calculate longitudinal changes. Pearson correlation was used for the correlation analysis.

**Feature volume of interests based analysis**

To compare CNN-based biomarker with conventional quantification methods, feature volume of interests (VOIs)-based analyses for FDG and AV-45 PET were carried out. The images were processed and quantified by feature VOI as described previously [16-18]. In brief, FDT PET volumes were spatially normalized with nonlinear warping process. Average FDG uptake from the angular (right/left), temporal (right/left) and posterior cingulate cortices were calculated. The quantification data were expressed relative to the mean uptake of a reference regions, pons and cerebellar vermis [16]. For AV-45 PET, cortical uptake of AV-45 was calculated. Cortical regions were segmented with FreeSurfer, and used to extract mean AV-45 uptake in frontal, anterior and posterior cingulate, lateral parietal, and lateral temporal regions. The overall cortical mean uptake value expressed relative to uptake in the whole cerebellum was used [17, 18].

**Statistics**

We compared the diagnostic and prediction accuracy of CNN with those of feature VOI-based approach with McNemar's nonparametric test. ROC analysis with area under the curve (AUC) measurement was performed for ConvScore and feature VOI-based parameters. The AUCs were compared using a nonparametric test of DeLong for comparison of two correlated ROC curves [19]. ConvScore was correlated with longitudinal changes of cognitive measurements using Pearson's correlation. All statistical analysis was performed by using the MATLAB software. P-value < 0.05 was considered significant.

# Results

## Results of AD classification and MCI conversion prediction

We included a total of 492 subjects in this study. Among them, 139 patients were AD, 171 patients were MCI and 182 were NC. Demographic data and cognitive measurements of each group were summarized in **table 1**.

The classification accuracy of the CNN-based approach was compared with feature VOI-based analysis of FDG and AV-45 PET as conventional quantitative analyses. Sensitivity, specificity and accuracy of CNN-based approach for classification between AD and NC were 93.5%, 97.8% and 96.0%, respectively. They were significantly higher than those of VOI-based analyses (*p*-values were summarized in **table 2**). Those for the prediction of MCI conversion was 81.0%, 87.0% and 84.2%, respectively. Specificity of CNN-based approach was significantly higher than VOI-based AV-45 PET. Sensitivity and accuracy were also higher compared with conventional methods, though it did not reach statistical significance (*p*-values were summarized in **table 2**).

In addition to the group classification, we calculated quantitative score, ConvScore, for predicting MCI converters. It was obtained from the value of the last layer of the network. Using ConvScore, ROC curves were drawn and AUC were calculated (**figure 2, table 2**). AUC of ConvScore was significantly higher than feature-VOI based analysis for AD classification ($p < 0.001$ for deep CNN vs. FDG and vs. AV-45) and prediction of MCI conversion ($p < 0.01$ for deep CNN vs. FDG and $p < 0.05$ for deep CNN vs. AV-45).

**Table 1.** Demographic data

| Variables | Normal (n=182) | AD (n=139) | p | MCI converter (n=79) | MCI nonconverter (n=92) | p |
|---|---|---|---|---|---|---|
| Age (years) | 73.4±6.3 | 74.3±8.2 | n.s. | 72.3±7.2 | 70.3±6.3 | n.s. |
| Sex (M/F) | 88/94 | 80/59 | n.s. | 43/36 | 51/41 | n.s. |
| Education (years) | 16.6±2.5 | 15.8±2.7 | <0.01 | 16.3±2.6 | 16.4±2.7 | n.s. |
| MMSE | 29.0±1.2 | 23.0±2.1 | <0.001 | 27.1±1.9 | 28.1±1.7 | <0.001 |
| ADAS | 9.1±4.5 | 31.0±8.6 | <0.001 | 22.0±6.8 | 12.9±5.5 | <0.001 |
| APOE ε4 (-/+) | 130/52 (28.6%) | 44/95 (68.3%) | <0.001 | 22/57 (72.2%) | 58/34 (37.0%) | <0.001 |

**Table 2.** Sensitivity, specificity and accuracy for the discrimination between AD and normal controls and the prediction for MCI converters.

| | Feature VOI-based | | Deep learning-based approach | p-value | |
|---|---|---|---|---|---|
| | FDG | AV45 | | vs. FDG | vs. AV45 |
| **AD vs. Normal** | | | | | |
| Sensitivity,% | 84.2 | 81.8 | 93.5 | 0.004 | <0.001 |
| Specificity,% | 86.3 | 80.8 | 97.8 | <0.001 | <0.001 |
| Accuracy,% | 85.4 | 80.7 | 96.0 | <0.001 | <0.001 |
| ROC AUC | 0.91 | 0.84 | 0.98 | <0.001 | <0.001 |
| **MCI Conversion Prediction** | | | | | |
| Sensitivity,% | 70.9 | 86.1 | 81.0 | 0.08 | 0.29 |
| Specificity,% | 79.3 | 75.0 | 87.0 | 0.19 | 0.007 |
| Accuracy,% | 75.4 | 80.1 | 84.2 | 0.02 | 0.21 |
| ROC AUC | 0.82 | 0.83 | 0.89 | 0.006 | 0.03 |

ROC: receiver operating characteristic; AUC: area under curve

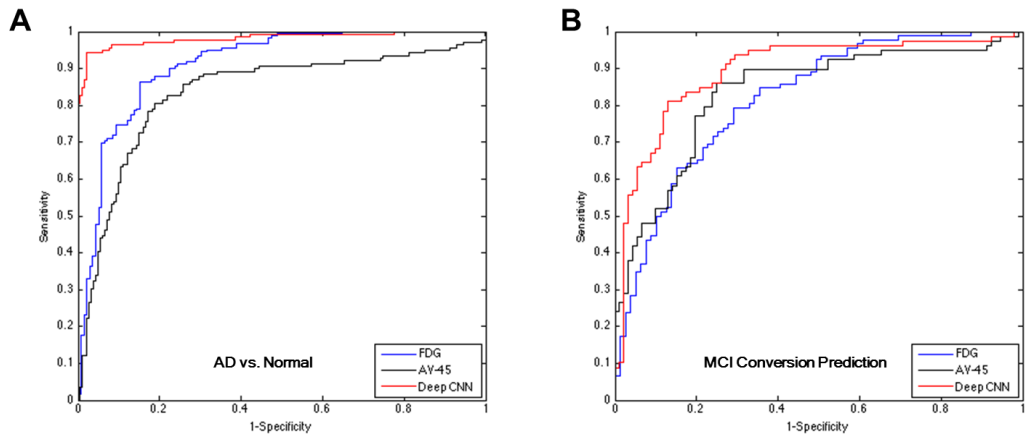

**Figure 2** ROC curves for deep CNN. ROC analyses were performed for the classification of AD (A) and the prediction of MCI conversion (B). ROC curves of feature VOI-based approaches using FDG and AV-45 PET were also drawn. Areas under curve (AUC) values were calculated. (A) AUC of ConvScore was significantly higher than feature-VOI based analysis for AD classification (AUC=0.98, 0.91 and 0.84 for ConvScore, FDG and AV-45, respectively; $p < 0.001$ for deep CNN vs. FDG and vs. AV-45) (B) AUC of ConvScore was also significantly higher than feature VOI-based analysis for prediction of MCI conversion (AUC=0.89, 0.82 and 0.83 for ConvScore, FDG and AV-45, respectively; $p < 0.01$ for deep CNN vs. FDG and $p < 0.05$ for deep CNN vs. AV-45).

**Correlation of CNN-based biomarker with cognitive outcomes**

ConvScore calculated from baseline PET images of MCI patients was significantly correlated with the longitudinal change of cognitive measurements at 1 year and 3 years (**figure 3**). ConvScore was significantly positively correlated with longitudinal change of CDR-SB (r=0.37, p<0.0001 at 1 year and r=0.63, p<0.0001 at 3 years), ADAS-Cog (r=0.29, p=0.0001 at 1 year and r=0.24, p=0.004 at 3 year) and FAQ (r=0.40, p<0.0001 at 1 year and r=0.67, p<0.0001 at 3 year). It was significantly negatively correlated with MMSE (r=-0.30, p<0.0001 at 1 year and r=-0.61, p<0.0001 at 3 year). Note that the changes were steeper for the measurements at 3 years compared with 1 year. ConvScore of MCI converters was significantly higher than that of MCI nonconverters (2.40±1.49 and -0.13±1.36, respectively. p<0.0001) (**figure 4**). ConvScore could be used as a quantitative biomarker for predicting longitudinal cognitive measurements decline in MCI patients as well as conversion to AD.

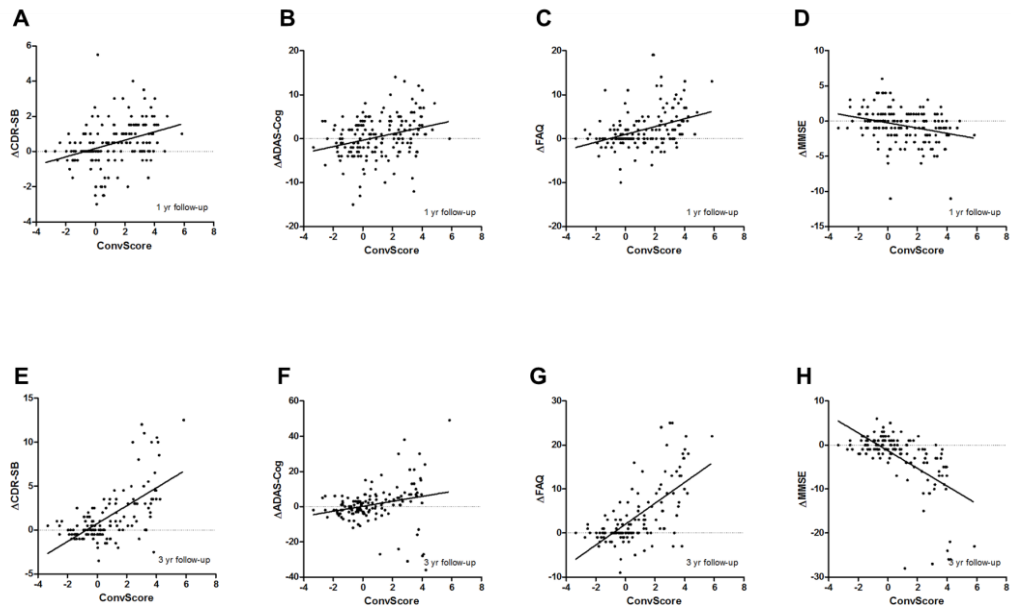

**Figure 3** Correlation between output of the network and longitudinal changes of cognitive measurements. The last layer provides an output scores for AD or MCI converter, which are defined as ConvScore. Cognitive measurements at baseline, 1 year, and 3 years were obtained for each MCI patient and ConvScore was correlated with the longitudinal change of them. (A-D) ConvScore was significantly correlated with the change of Cognitive Dementia Scaling Sum of Boxes (CDR-SB) (r=0.37, p<0.0001), Alzheimer Disease Assessment Scale-Cognitive Subtest (ADAS-Cog) (r=0.29, p=0.0001), Functional Activities Questionnaire (FAQ) (r=0.40, p<0.0001) and Mini-mental State Examination (MMSE) (r=-0.30, p<0.0001) from baseline to 1 year follow-up. (E-H) The significant correlation between ConvScore and the change of the measurements from baseline to 3 years was also found (r=0.63, p<0.0001 for CDR-SB; r=0.24, p=0.004 for ADAS-Cog; r=0.67, p<0.0001 for FAQ; r=-0.61, p<0.0001 for MMSE).

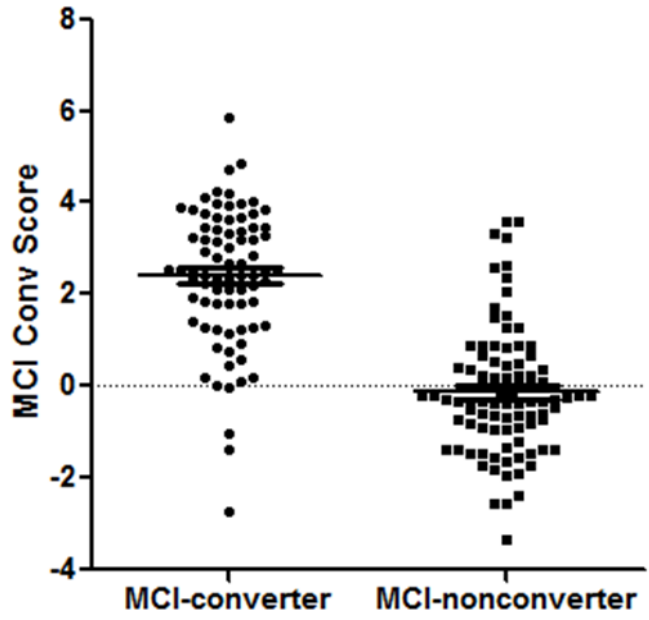

**Figure 4** ConvScore of MCI converters and nonconverters. MCI converters were significantly higher

ConvScore than nonconverters (2.40±1.49 and -0.13±1.36, respectively. p<0.0001).

## Discussion

In the present study, we developed a deep learning-based diagnostic method for predicting cognitive decline in MCI patients. According to our knowledge, we firstly applied recent deep CNN to multimodal PET images to predict future outcome. Deep CNN could accurately classify patients' diagnostic group with the minimal steps of image processing and provide a quantitative biomarker for predicting cognitive outcome. Accordingly, the prediction of cognitive decline in our method could be automatically made by simply inputting subjects' images. Output value (ConvScore) was significantly correlated with the longitudinal change of cognitive measurements.

Previous PET imaging biomarkers relied on the uptake value in a set of regions of interest developed *a priori* or whole-cortical uptake value [6, 9, 10]. To obtain these values, several image processing steps such as spatial normalization and cortical segmentation using structural MRI were required. However, such processing was not standardized and nonlinear image transformation could introduce a potential source of errors particularly in morphological alterations [20, 21]. Our proposed method used spatially unnormalized baseline image data of AD and NC. It suggests that our approach is simply able to be utilized to redesign the voxelwise brain imaging processing pipeline which has routinely implemented normalization to template space. Furthermore, accuracy of differentiation between MCI converter and nonconverter (84.2%) outperformed the conventional feature-VOI methods to discriminate MCI converters. ROC comparison results also revealed that the accuracy of prediction was significantly higher than other methods. It is the benefit of deep CNN that could automatically discover the optimal features for image classification [11]. Furthermore, the accuracy also outperformed other state-of-the-art machine-learning algorithms based on feature selection from multiple images for this differentiation problem [22-25], though those studies used different imaging modalities and clinical variables.

Our approach could provide a quantitative variable, ConvScore, to be used as a fusion biomarker for multiple PET images. As an imaging biomarker, low glucose metabolism and high amyloid deposit in the cortex at baseline can predict the longitudinal decline of cognitive scores [16, 17]. However, a combined parameter considering both metabolism and amyloid deposit has been needed. ConvScore is not only a fusion biomarker directly obtained by both PET images, but strongly correlated with longitudinal cognitive measurements. It suggests cognitive functions of patients with high ConvScore at baseline could be rapidly deteriorated. This strong correlation is an important observation that has impact on clinical trials for early treatment intervention in prodromal AD. ConvScore could help select the subjects who would benefit from treatment in the clinical trials.

In the clinical setting, most imaging studies are assessed by experts' visual analysis, because it is simpler and more practical than the quantitative assessment which needs time-consuming procedures. Deep CNN is motivated by human visual perception, which hierarchically processes recognized images in the cerebral cortex [26]. As hierarchical features of images are automatically trained by data, manual feature selection or image processing steps can be minimized in deep CNN [11, 26]. Therefore, after the training, the network is automatically able to analyze patients' images by simply inputting subjects' images. Considering the ease of application, the CNN-based image interpretation system has a large potential to be used in development of biomarkers of several diseases including cancer, cardiovascular disorders as well as neurodegenerative diseases.

Recent remedies of CNN for achieving higher accuracy are increasing the depth of the network [27]. However, to learn a deeper neural network, more image data will be essential. As the network is trained by the larger data, the higher performance it shows. In our study, to overcome the limited number of imaging data, PET images were augmented by flipping image in left-right direction. It was based on the previous knowledge that AD and MCI converters showed symmetrically decreased FDG uptake and increased AV-45 uptake in the cerebral cortices. The network trained without this augmentation process showed 89.4% and 81.3% accuracy for the differentiation between AD and NC and predicting MCI converters, respectively. Though the augmentation process increased the performance of the network, it might cause potential error in the classification because the two brain hemispheres have partly different functions. In the future, larger image data cohort and deeper network architecture could improve the CNN-based approach.

Our deep CNN-based approach could predict cognitive decline in MCI patients with very high accuracy. For testing whether a MCI subject would convert to AD, baseline PET images without spatial transformation were needed as a feature extraction was automatically performed in multiple layers of the network. As a future work, our approach may be additionally used in completely independent cohorts. In addition, as an accurate biomarker, it could help select appropriate prodromal patients who benefit from early intervention.

# Acknowledgement

Data collection and sharing for this project was funded by the Alzheimer's Disease Neuroimaging Initiative (ADNI) (National Institutes of Health Grant U01 AG024904) and DOD ADNI (Department of Defense award number W81XWH-12-2-0012). ADNI is funded by the National Institute on Aging, the National Institute of Biomedical Imaging and Bioengineering, and through generous contributions from the following: AbbVie, Alzheimer's Association; Alzheimer's Drug Discovery Foundation; Araclon Biotech; BioClinica, Inc.; Biogen; Bristol-Myers Squibb Company; CereSpir, Inc.;Eisai Inc.; Elan Pharmaceuticals, Inc.; Eli Lilly and Company; EuroImmun; F. Hoffmann-La Roche Ltd and its affiliated company Genentech, Inc.; Fujirebio; GE Healthcare; IXICO Ltd.; Janssen Alzheimer Immunotherapy Research & Development, LLC.; Johnson & Johnson Pharmaceutical Research & Development LLC.; Lumosity; Lundbeck; Merck & Co., Inc.;   Meso Scale Diagnostics, LLC.; NeuroRx Research; Neurotrack Technologies; Novartis Pharmaceuticals Corporation; Pfizer Inc.; Piramal Imaging; Servier; Takeda Pharmaceutical Company; and Transition Therapeutics. The Canadian Institutes of Health Research is providing funds to support ADNI clinical sites in Canada. Private sector contributions are facilitated by the Foundation for the National Institutes of Health (www.fnih.org). The grantee organization is the Northern California Institute for Research and Education, and the study is coordinated by the Alzheimer's Disease Cooperative Study at the University of California, San Diego. ADNI data are disseminated by the Laboratory for Neuro Imaging at the University of Southern California.

# Disclosure

None declared.